%% file: cvpr.tex
\documentclass[final]{cvpr}

\usepackage{times}
\usepackage{epsfig}
\usepackage{graphicx}
\usepackage{amsmath}
\usepackage{amssymb}
\usepackage{booktabs}
\usepackage{multirow}
\usepackage{graphics}
\usepackage{subfigure}
\usepackage[pagebackref=false,breaklinks=true,colorlinks,bookmarks=false]{hyperref}

\newcommand{\tabincell}[2]{\begin{tabular}{@{}#1@{}}#2\end{tabular}}

\def\confYear{CVPR 2021}

\begin{document}

%%%%%%%%% TITLE
\title{Inception Convolution with Efficient Dilation Search}

% \and
\author{
    Jie Liu\textsuperscript{1}$^{\ast}$, Chuming Li\textsuperscript{2}$^{\ast}$, Feng Liang\textsuperscript{2,5}, Chen Lin\textsuperscript{3}, Ming Sun\textsuperscript{2,5}, Junjie Yan\textsuperscript{2}, Wanli Ouyang\textsuperscript{4}, Dong Xu\textsuperscript{4}\\
    \textsuperscript{1}{Beihang University}, \textsuperscript{2}{SenseTime Research} \\
    \textsuperscript{3}University of Oxford, \textsuperscript{4}The University of Sydney, \textsuperscript{5}Shanghai Al Laboratory \\
    \small \tt ljie@buaa.edu.cn, \small \tt chen.lin@eng.ox.ac.uk, \small \tt \{wanli.ouyang, dong.xu\}@sydney.edu.au \\
    \small \tt \{lichuming, liangfeng, sunming1, yanjunjie\}@sensetime.com 
}

\maketitle
\pagestyle{empty}
\thispagestyle{empty}
%%%%%%%%% ABSTRACT
\input{sections/0.abstract}

%%%%%%%%% BODY TEXT
\input{sections/1.introduction}

\input{sections/2.relatedwork}

\input{sections/3.method}

\input{sections/4.experiment}

\input{sections/5.conclusion}
%-------------------------------------------------------------------------

\clearpage
{\small
\bibliographystyle{ieee_fullname}
\bibliography{egbib}
}
% \clearpage
% \appendix
% \input{sections/supplementary}
\end{document}

%% file: sections/0.abstract.tex
\begin{abstract}
{\let\thefootnote\relax\footnote{{$^{*}$Equal contribution}}}As a variant of standard convolution, a dilated convolution can control effective receptive fields and handle large scale variance of objects without introducing additional computational costs. 
To fully explore the potential of dilated convolution, we proposed a new type of dilated convolution (referred to as inception convolution), where the convolution operations have independent dilation patterns among different axes, channels and layers. To develop a practical method for learning complex inception convolution based on the data, a simple but effective search algorithm, referred to as efficient dilation optimization (EDO), is developed. Based on statistical optimization, the EDO method operates in a low-cost manner and is extremely fast when it is applied on large scale datasets. Empirical results validate that our method achieves consistent performance gains for image recognition, object detection, instance segmentation, human detection, and human pose estimation.
For instance, by simply replacing the 3 $\times$ 3 standard convolution in the ResNet-50 backbone with inception convolution, we significantly improve the AP of Faster R-CNN from 36.4\% to 39.2\% on MS COCO.

\end{abstract}

%% file: sections/1.introduction.tex
\section{Introduction}
\label{introduction}
As an important concept of convolution neural network, the receptive field has been extensively studied. In \cite{luo2016understanding}, Luo~\etal showed that the intensity in each receptive field roughly obeys a Gaussian distribution and only few pixels around the central part of the receptive field can effectively contribute to the response of the output neuron. Furthermore, in the previous works \cite{liang2019computation, peng2019efficient}, a more carefully defined optimal receptive field (ORF) has been evaluated for different tasks.

\begin{figure}[] %[H]
    \centering 
    \includegraphics[width=0.35\textwidth]{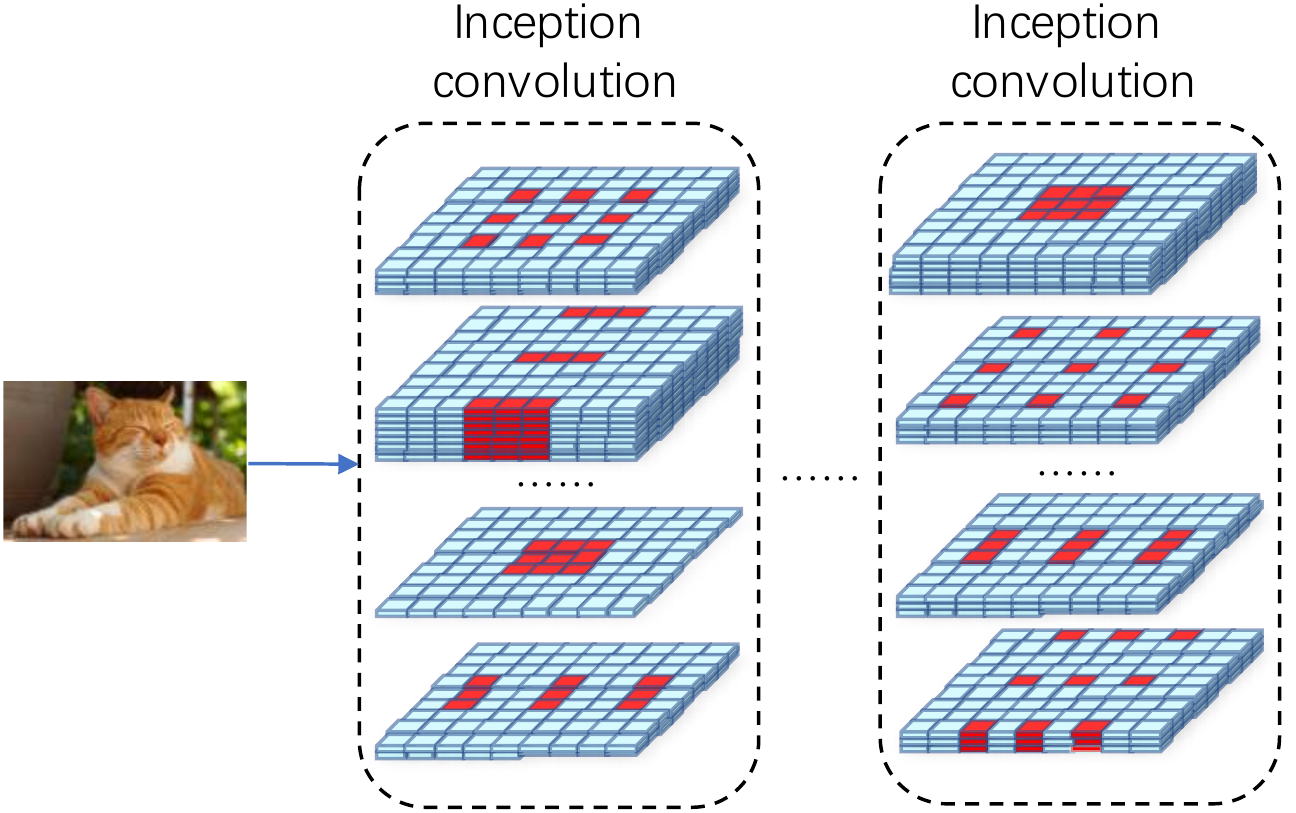}
    \caption{Inception convolution contains rich dilation patterns along both spatial axes and channels in each convolution layer.}
    \label{fig:first_fig_inception_conv}
\end{figure}

The requirement of the optimal receptive field is due to the variance of the input image sizes or the scales of objects of interest. For instance, for image classification, the input sizes tend to be small (e.g, 224 $\times$ 224), while for object detection, the input sizes are much larger and the objects can be from a large range of scales. Correspondingly, different tasks would require different ERFs. The different requirements of ERFs from different tasks makes it necessary to develop a general and practical optimization algorithm to learn the optimal ERF for a specific task. 

As discussed in \cite{luo2016understanding}, the dilation value of dilated convolution kernels is a highly effective hyper-parameter to control the receptive field for different tasks. The work in \cite{liang2019computation} proposed to assign different dilation values at different stages of a CNN, which achieves consistent performance improvements. Subsequently, NATS \cite{peng2019efficient} divided a convolution operation into different groups with each having independent dilation values. However, they apply the standard network architecture search methods in the relatively coarse search spaces, which neglects the fine-grained inner structure of dilated convolution. Therefore, in this work, we focus on exploring the search problem in the dilation domain to efficiently learn the optimal receptive field. 

First of all, we would like to have a more flexible search space when compared with \cite{liang2019computation}. Flexibility can make us learn the optimal receptive field in order to better fit to different datasets. As shown in Figure~\ref{fig:first_fig_inception_conv}, we propose a new type of dilated convolution, called Inception Convolution, which contains as much as possible dilation patterns. In the space of inception convolution, the dilation pattern along each axis, each channel, and each convolution layer is independently defined. As a result, a dense range of possible receptive fields are considered in our inception convolution.

For optimization, a direct solution is to use the existing works in neural architecture search (NAS), which enables automatic search for the optimal combination of various network operations. DARTS~\cite{liu2018darts} and single path one-shot~\cite{guo2020single} (SPOS) are two main families of efficient NAS methods. DARTS trained a supernet, where discrete operation selection was relaxed to a continuous weighted sum of the output from all candidate operations. After training, in each block, the operation with the largest architecture weight was chosen. SPOS randomly selected an operation sequence (subnet) from a pre-trained supernet and the same operation in different sequences share the same weights. After training, SPOS selected the best operation sequence via sampling and evaluation of multiple sequences with the shared-weights. 

However, both DARTS and SPOS are not suitable for our huge search space. In DARTS, during training all operations in a block are applied to the input to make the architecture weights aware of each operation's importance, but the number of dilation patterns for a convolution layer (block) is large, \ie, 16 if each of the two axes has 4 choices. It means DARTS requires 16 sequential calculations, thus it has low GPU utility and huge computational costs. SPOS samples the operation sequences during training. However, in our search space, the number of dilation patterns even in a single convolution layer is huge, \ie, $(d_{max})^{2C}$, where $C$ is the number of channels  and $d_{max}$ is the maximum dilation value. Due to the huge number of dilation patterns (paths), it is an extremely difficult task for SPOS as well.

In this work, we propose a simple and efficient dilation optimization algorithm (EDO). In EDO, each layer of the supernet is a standard convolution operation whose kernel covers all possible dilation patterns. After pre-training of the supernet, we select the dilation pattern for each channel in each convolution layer by solving a statistical optimization problem. Specifically, for each layer, according to the pre-trained weights, we minimize the $L_1$ error between the output of the original convolution layer and the output of the learned dilated convolution layer with the selected dilation pattern, based on which we can learn the optimal dilation pattern for this layer. 

EDO supports efficient channel-wise dilation pattern selection over our complete dilation pattern search space. When compared with the search based method in \cite{guo2020single}, the search cost of our methods is very low. When compared with the differentiable NAS methods \cite{liu2018darts,cai2018proxylessnas}, EDO converts sequential calculation related to different dilation patterns into a parallel way, thus it has lower computation cost and higher GPU utilization. 
Further, when compared with SPOS, we do not need to design further mechanism to handle the extremely large number of dilation patterns (paths).

Our contributions are three folds: 
1) We propose a new type of dilated convolution, referred to as inception convolution, which can be readily combined with a large number of backbones for various visual recognition tasks;
2) We also propose a low-cost statistical optimization based network architecture search algorithm EDO, which can efficiently learn the optimal receptive field based on the data; 
3) Comprehensive experiments demonstrate that our learnt inception convolution in combination with various backbones generally achieves performance improvement for various visual tasks without introducing any additional computational costs.

\begin{figure*}[h] %[H]
    \centering 
    \includegraphics[width=0.72\textwidth]{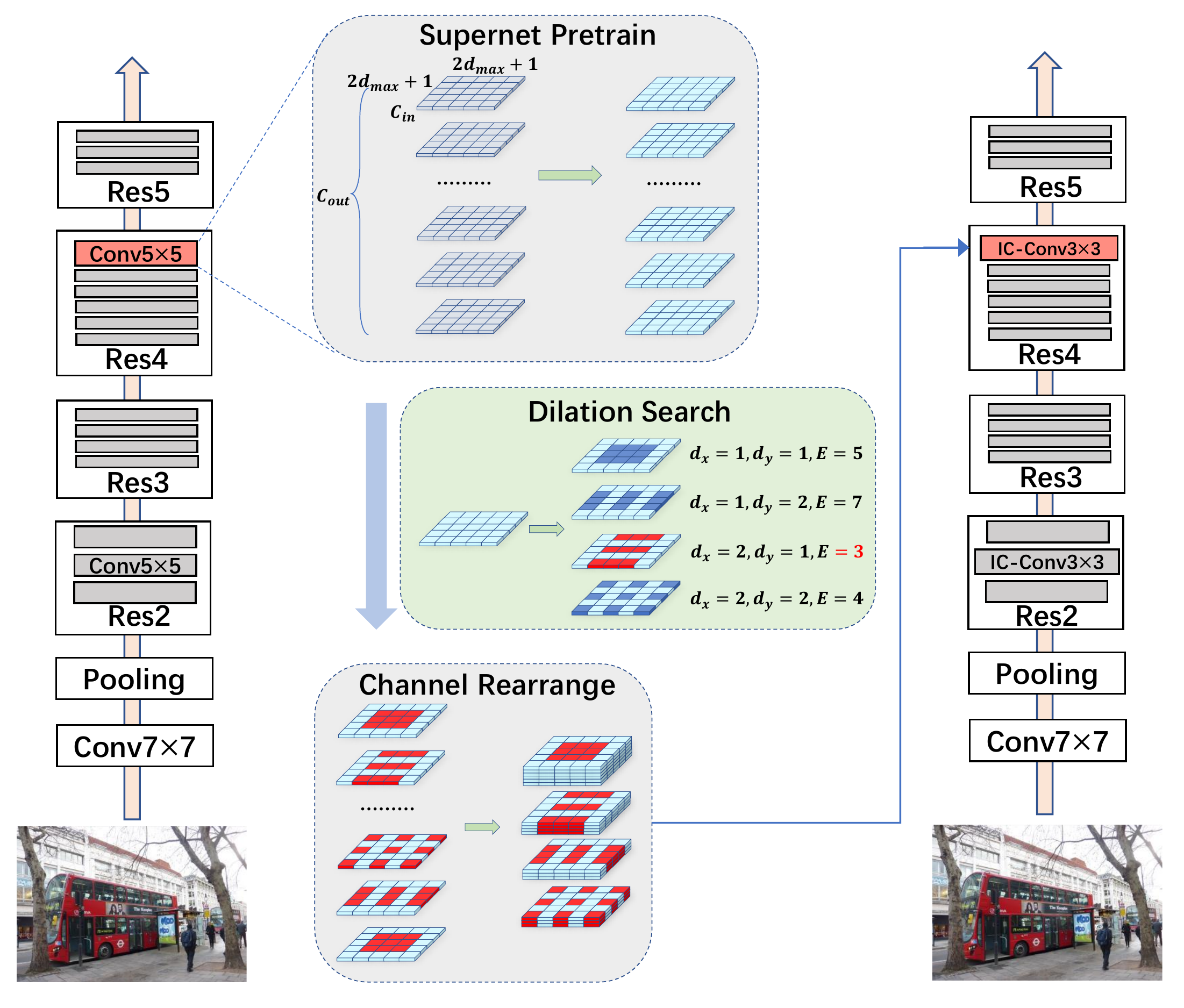}
    \caption{An overview of our EDO algorithm. ResNet50 (R50) is taken as an example. Firstly, for the convolution layers in R50 with the kernel size of $3\times 3$, we change the kernel size to $(2d_{max} + 1) \times (2d_{max} + 1)$, where we set $d_{max}$ = 2 in this figure, namely the kernel size of the supernet is $5\times 5$. Secondly, we select the optimal dilation pattern with the best representation ability, which leads to the minimum representation error $E$ (see Section~\ref{solution} for more details). For one channel, dilation (2,1) with $E$ = 3 is selected in this example. Finally, we rearrange the filters so that the filters with the same dilation patterns are arranged together, which produce our Inception Convolution.}
    \label{fig:pipeline}
\end{figure*}

%% file: sections/2.relatedwork.tex
\section{Related Work}
\label{relatedwork}

\subsection{Receptive Field}

The Receptive Field is a crucial concept in Convolutional Neural Networks (CNNs). Traditional CNNs~\cite{krizhevsky2017imagenet, simonyan2014very, he2016deep} stacked multiple convolutional layers to enhance the receptive field. The inception networks~\cite{szegedy2015going, szegedy2016rethinking, szegedy2016inception} introduced different size of filters, which are operated on the same level to aggregate different receptive fields.
 
Deformable convolution~\cite{dai2017deformable} predicted the sampling offsets with respect to the preceding feature maps to adjust the receptive fields automatically. Scale-Adaptive Convolutions~\cite{zhang2017scale} predicted local flexible-sized dilations at each position to cover objects of various sizes. However, these methods are unfriendly to hardware optimization and thus cannot be used for real-time applications.

Dilated (Atrous) convolution~\cite{yu2015multi, chen2017rethinking} changes the receptive field by performing convolution at sparsely sampled locations, which have been widely used in semantic segmentation~\cite{chen2017deeplab, zhao2017pyramid} and object detection~\cite{li2018detnet, li2019scale, peng2019pod}. PSConv~\cite{li2020psconv} manually mixed up a spectrum of dilation rates in one convolution, which is shown to be sub-optimal in our experiments. In this work, we aim at searching for efficient inception convolution, which can be treated as a mixed dilated convolution and it is also friendly to hardware optimization.

\subsection{Neural Architecture Search}
\vspace{-2mm}
Neural architecture search has attracted increasing attention. Early NAS approaches~\cite{zoph2016neural, zoph2018learning} are computationally expensive due to the evaluation
of each candidate. To reduce the searching costs, EcoNAS~\cite{zhou2020econas} proposed an EA-based algorithm to improve searching efficiency. Recently, One-shot NAS methods~\cite{bender2018understanding, liu2018darts, cai2018proxylessnas, guo2020single, cai2019once,li2020improving,guo2020powering,liu2020block} built a direct acyclic graph $G$ (a.k.a., supernet) to subsume all architectures in the search space to further reduce the costs. More relevant to our work, some NAS works were proposed to search for dilated convolutions. NATS~\cite{peng2019efficient} employed the DARTS method~\cite{liu2018darts} to search for the dilated rate at the group level in the CNN backbone. CRNAS~\cite{liang2019computation} searched for different dilation rates for different building blocks by using the SPOS method~\cite{guo2020single}. However, the search spaces of the aforementioned methods are limited. Moreover, simply adopting existing searching method, like DARTS~\cite{liu2018darts} or SPOS~\cite{guo2020single} cannot work well for our search task as these methods cannot handle the extremely large number of possible operations in a single convolution layer.

%% file: sections/3.method.tex
\section{Methodology}
\label{method}

\subsection{Problem Formulation of Inception Convolution}
To fully explore the flexibility of dilation, in our inception convolution we consider a complete dilation space. An Inception Convolution has independent dilation values for the two axes in each channel and is formally represented as follows:

\begin{equation}
\begin{aligned}
    \boldsymbol{d} = & \{ ({d}_x^i,{d}_y^i) | {d}_x^i \in {\{1,2,..,d_{max}\}}, {d}_y^i \in {\{1,2,..,d_{max}\}}, \\
    & i \in \{1,2,...,C^{out}\} \}, 
    \label{dilation_domain}
\end{aligned}
\end{equation}
where ${d}_x^i$ and ${d}_y^i$ are the dilation values in $x$ axis and $y$ axis of the filter at the $i$-th output channel, ranging from $1$ to $d_{max}$, and $C^{out}$ denotes the number of output channels. The total number of all possible dilation patterns in a single inception convolution is ${(d_{max})^{2C_{out}}}$. In this work, we aim to develop an algorithm to efficiently learn the optimal receptive field for different tasks by selecting the optimal $\boldsymbol{d}$.

\subsection{Solution}
\label{solution}
Recently, NAS evolves as an effective way to learn high-performance network architectures in the specified search spaces. DARTS and SPOS are two main-stream families of NAS methods. However, as inception convolution contains $(d_{max})^2$ dilation patterns and $(d_{max})^{2C_{out}}$ candidates, both DARTS and SPOS cannot be used for efficient search in the huge search space.

Both works SPOS and DARTS show that the weights in a pre-trained supernet are informative to guide the selection operation. In this work, we follow this idea and formulate a statistical optimization problem to select the optimal dilation patterns based on the corresponding weights in a pre-trained supernet. Our optimization algorithm, EDO, is simple, effective, and efficient. The pipeline of our proposed method is shown in Figure~\ref{fig:pipeline}.

\textbf{Supernet.} Given a network architecture, we construct a supernet and then we keep the architecture unchanged but change the kernel sizes to cover all candidate dilation patterns. Formally, for a convolution layer in the supernet with the kernel size of $2k+1$, we replace it  with the convolution kernel with the size of $2kd_{max}+1$, which is the maximum width and height for all candidate dilation patterns. Note that the supernet is pre-trained for each given task. 

\textbf{Statistical Optimization.} For each convolution layer with the weights $\boldsymbol{W}$, we define $\boldsymbol{W}^i$ as the weights of the $i$-th convolutional filter in our supernet, and denote $\boldsymbol{d}$ as the set of $\{({d}_x^i,{d}_y^i)|_{i=1}^{C_{out}}\}$ where ${d}_x^i$ and ${d}_y^i$ represent the sampled positions from the $i$-th channel. Based on ${d}_x^i$ and ${d}_y^i$, we can generate $\boldsymbol{\widetilde{W}}_{{d}_x^i, {d}_y^i}^i$, in which the sampled positions are filled with the corresponding values from $\boldsymbol{W^i}$ while the unsampled positions are filled with zeros. We stack $\boldsymbol{\widetilde{W}}_{{d}_x^i, {d}_y^i}^i$ along the output channel dimension and produce $\boldsymbol{\widetilde{W}}_{\boldsymbol{d}}$. 
 Note the dimensions of $\boldsymbol{W}$ and $\boldsymbol{\widetilde{W}}_{\boldsymbol{d}}$ are the same (\ie, 
 $\boldsymbol{W},\boldsymbol{\widetilde{W}}_{\boldsymbol{d}}\in \mathbb{R}^{C_{out} \times C_{in} \times (2kd_{max}+1) \times (2kd_{max}+1)}$) and the dimensions of $\boldsymbol{W}^i$ and $\boldsymbol{\widetilde{W}}_{{d}_x^i, {d}_y^i}^i$ are also the same (\ie, $\boldsymbol{W}^i, \boldsymbol{\widetilde{W}}_{{d}_x^i, {d}_y^i}^i \in \mathbb{R}^{C_{in} \times (2kd_{max}+1) \times (2kd_{max}+1)}$), where $C_{in}$ and $C_{out}$ are the numbers of input and output channels.
We formulate the dilation pattern selection task as an optimization problem,
where the $L_1$ error between the expectation of the output from the pre-trained  weights $\boldsymbol{W}$ and the expectation of the output from the sampled dilation weights $\boldsymbol{\widetilde{W}}_{\boldsymbol{d}}$ is minimized.
Formally,
we arrive at:
\begin{equation} 
\label{eq:Opt}
\begin{aligned}
    \mathop{\min}_{\boldsymbol{d}} \ \ & \|E[\boldsymbol{W}\ast \boldsymbol{X}] - E[\boldsymbol{\widetilde{W}}_{\boldsymbol{d}}\ast \boldsymbol{X}]\|_1, \\
    s.t. \ \ & {d}_x^i \in \{1,...,d_{max}\}, {d}_y^i \in \{1,...,d_{max}\}, 
\end{aligned}
\end{equation}
where $\boldsymbol{X} \in \mathbb{R}^{B \times C_{in} \times H \times W } $ is the input of this convolution layer with the batch size of $B$, the number of input channels $C_{in}$, the height $H$ and the width $W$. We use $\ast$ to denote the convolution operator, and $E$ is the expectation operator. As $\boldsymbol{W}$ and $\boldsymbol{\widetilde{W}}_{\boldsymbol{d}}$ are independent of $\boldsymbol{X}$, the  objective function in Eq.~(\ref{eq:Opt}) is further expressed as follows:

\begin{equation} 
\begin{aligned}
    & \|E[\boldsymbol{W}\ast \boldsymbol{X}] - E[\boldsymbol{\widetilde{W}}_{\boldsymbol{d}}\ast \boldsymbol{X}]\|_1 \\
    &= \|\boldsymbol{W} \ast E[\boldsymbol{X}] - \boldsymbol{\widetilde{W}}_{\boldsymbol{d}} \ast E[\boldsymbol{X}]\|_1, \\
    &= \| (\boldsymbol{W} - \boldsymbol{\widetilde{W}}_{\boldsymbol{d}})\ast E[\boldsymbol{X}]\|_1.
\end{aligned}
\end{equation}

In the above optimization problem, we assume that, after batch normalization, the mean feature value does not vary too much across different positions of $\boldsymbol{X}$, which means that $E[\boldsymbol{X}]$ has almost the same value among all positions. Assuming the mean value is $\alpha$, then we propose to solve an alternative optimization problem as follows:

\begin{equation} 
\begin{aligned}
   \mathop{\min}_{\boldsymbol{d}} \sum_{i=1}^{C^{out}} \alpha \| (\boldsymbol{W}^i - \boldsymbol{\widetilde{W}}_{{d}_x^i, {d}_y^i}^i) \ast \boldsymbol{1} \|_1, \label{final_target}
\end{aligned}
\end{equation}

where $\boldsymbol{1}$ is an all-ones matrix with the same dimension as $\boldsymbol{X}$, and $\alpha$ is a constant scalar which can be omitted in optimization. Based on Eq.~(\ref{final_target}), the optimal $\boldsymbol{d}$ of each convolution layer can be easily solved, by independently traversing all dilation patterns $({d}_x^i, {d}_y^i)$ for each filter $\boldsymbol{W}^i$ with little cost.

\subsection{Discussion}
\textbf{Relationship with DARTS.} An intuitive application of DARTS to our inception convolution is introduced in \cite{peng2019efficient}. In DARTS, $(d_{max})^2$ operations are calculated sequentially, while our EDO algorithm can calculate the $(d_{max})^2$ operations in parallel. Besides, the total cost of DARTS is  $C_{in} \times  C_{out} \times (2k+1)^2 \times (d_{max})^2$,
but the total cost of our EDO is $C_{in} \times C_{out} \times (2kd_{max}+1)^2$. For most CNNs, where $k$ is usually 1, $C_{in} \times C_{out} \times (2kd_{max}+1)^2$ is only 56\% of $C_{in} \times C_{out} \times (2k+1)^2 \times (d_{max})^2$ when $d_{max}$ equals to 4. Therefore, EDO is more efficient than DARTS.

Additionally, as shown in \cite{zela2019understanding}, DARTS degenerates into random sampling in some cases because the principal eigenvalue of the Hessian matrix of the architecture parameters appears large. However, we directly define the statistical optimization problem over the pre-trained network weights rather than introducing the architecture parameters,
which is more  robust than DARTS. 
\vspace{2mm}

\textbf{Relationship with NATS and CRNAS.} While NATS and CRNAS also considers the search space for searching flexible dilation patterns, it is less complete when compared with our inception convolution. CRNAS searches the dilation patterns independently at each stage, thus it is based on SPOS. NATS divides a convolution into the groups and search among a few dilation patterns (usually 5 patterns) for each group with DARTS.
In contrast, our inception convolution is channel-wise and contains all dilation patterns within the max dilation value $d_{max}$.

%% file: sections/4.experiment.tex
\section{Experiments}
\label{experiment}

\subsection{Image Recognition}
\subsubsection{Dataset and implementation details}
For image recognition, we evaluate our method on the ImageNet dataset~\cite{russakovsky2015imagenet} with 1.28M training images and 50k validation images. We first train our supernet with the largest kernel size (\ie, 9) and follow the standard training procedure in ~\cite{he2016deep}. More specifically, we use stochastic gradient descent (SGD) as the optimizer with the momentum of 0.9 and the weight decay of 0.0001. The supernet is trained for 100 epochs with the batch size of 1024 without any trick. We adopt the cosine learning rate scheduler with the initial learning rate of 0.4. Then we perform EDO to obtain the best inception convolutions, as described in Section~\ref{method}. The resultant IC-Net is retrained by using the same procedure as the supernet training process. Below, each method in combination with our inception convolution is also referred to as ``IC-X'' (e.g. ``IC-ResNet18'').

\subsubsection{Experimental results}
We search inception convolution based on various types of networks, from MobileNetV2~\cite{sandler2018mobilenetv2} to ResNeXt~\cite{xie2017aggregated}. As shown in Table~\ref{tab:imagenet}, our inception convolution consistently boosts the performance on ImageNet. In terms of top-1 accuracy, IC-ResNet18 and IC-ResNet50 outperform the baselines by 1.07\% and 1.11\%, respectively. Our inception convolution is also compatible with other networks consisting of depth-wise convolution or group convolution. For instance, in terms of top-1 accuracy, inception convolution leads to 1.21\% and 0.62\% gain on MobileNetV2 and ResNeXt101, respectively. 

\vspace{2mm}
\begin{table}[h]
	\caption{Top-1/top-5 accuracy (\%) comparison on the ImageNet validation set. All the methods are compared under the single center crop evaluation setting. \bf{The baseline results are re-implemented by ourselves based on the same code as our IC-Net.}}
	\label{tab:imagenet}
	\begin{center}
	    \small
		\begin{tabular}{c|c|c}
			\toprule
			Network Architectures & Conv Types & Top-1/5 Acc.(\%) \\
			\midrule
			\multirow{2}{*}{MobileNetV2~\cite{sandler2018mobilenetv2}}  & Standard & 70.71 / 89.81 \\
                             & IC-Conv & \bf{71.92 / 90.54} \\
			\hline
			\multirow{2}{*}{ResNet18~\cite{he2016deep}}  & Standard & 70.67	/ 89.74 \\
			                            & IC-Conv & \bf{71.74 / 90.91} \\
			\hline
			\multirow{2}{*}{ResNet50~\cite{he2016deep}}   & Standard & 76.19 / 92.93 \\
                                         & IC-Conv & \bf{77.30 / 93.58} \\
			\hline
			\multirow{2}{*}{ResNeXt101 (32x4d)~\cite{xie2017aggregated}} & Standard & 78.71 / 94.20 \\
                                                & IC-Conv & \bf{79.33 / 94.74} \\
			\bottomrule
		\end{tabular}
	\end{center}
\end{table}

\vspace{-2mm}

\subsection{Object Detection}
\label{sec:object_detection}
\subsubsection{Dataset and implementation details}
    \vspace{-2mm}
\label{detetion_training_cfg}
In the following experiments, unless otherwise stated, we will only replace the standard convolutions with our searched inception convolutions in the backbone.
For object detection, we use MS COCO 2017~\cite{lin2014microsoft} for the experiments. The dataset is challenging due to the huge variation of object scales and a large number of objects per image. The supernet with the largest kernel size (\ie, 9) is used as the pre-trained model to generate the inception convolutions. For detector training, we use stochastic gradient descent (SGD) as the optimizer with the momentum of 0.9 and the weight decay of 0.0001. The model is trained for 13 epochs, known as 1× schedule~\cite{Detectron2018}. We use the multi-GPU training strategy over eight 1080TI GPUs with a total batch size of 16. The initial learning rate is 0.00125 per image and is divided by 10 at the 8th and the 11th epochs. Warm-up is adopted for both baselines and our method.

\subsubsection{Experimental results}
Our searched inception convolution has great potential to be combined with various detectors, such as Faster R-CNN~\cite{ren2015faster} and Cascade R-CNN~\cite{cai2018cascade}. The same type of detectors is trained by using exactly the same 1× training procedure~\cite{Detectron2018}. We replace all the $3\times 3$ convolutions in the pre-trained backbone network by our inception convolution, while the convolutions in the FPN neck are kept as the standard convolutions.
As shown in Table~\ref{tab:detector}, our searched inception convolution boosts the performance of an extensive range of backbones on COCO. For Faster R-CNN~\cite{ren2015faster, lin2017feature} with FPN, our backbones of  IC-ResNet50, IC-ResNet101, and IC-ResNeXt101-32x4d outperform the baseline backbones of ResNet50, ResNet101, and ResNeXt101-32x4d by large margins of 2.5\%, 3.1\% and 1.6\% respectively. Our inception convolution is especially effective for large objects (4.1\%, 4.1\% and 2.4\% improvement
in terms of AP$_L$), possibly due to the large receptive field provided by dilation pattern search. For a more powerful method, Cascade R-CNN~\cite{cai2018cascade} with FPN, our method is also effective. Our backbone IC-ResNeXt101-32x4d achieves the AP of 45.7\%, which is 1.3\% higher than the baseline backbone.

\vspace{-2mm}
\subsubsection{Results from/on different detectors/datasets}

\begin{table}[h]
   \caption{Results (AP\%) of different detectors when using the standard convolution and our inception convolution on the COCO 2017 validation set. For all detectors, we achieve consistent performance improvements. For DETR, due to limited computational resources, we use the officially released training scripts with 150 epochs rather than the one with 500 epochs in the original work.}
   \label{tab:different_detector}   
      \begin{center}
      \resizebox{220pt}{!}{%
      \centering
      \begin{tabular}{c|cc|l}
      \toprule
      Detector                            & Backbone              & Conv Type & AP     \\ \hline
      \multirow{2}{*}{Faster R-CNN (C4)~\cite{ren2015faster}}     & \multirow{2}{*}{ResNet50}  & Standard  &35.0	  \\
                                          &                       & IC-Conv   &38.5$_{(\textbf{+3.5})}$     \\ \hline
      \multirow{2}{*}{RetinaNet~\cite{lin2017focal}}          & \multirow{2}{*}{ResNet50}  & Standard  &36.0	       \\
                                          &                       & IC-Conv   &37.9$_{(\textbf{+1.9})}$      \\ \hline
      \multirow{2}{*}{DETR~\cite{carion2020end}}               & \multirow{2}{*}{ResNet50}  & Standard  &39.7	    	         \\
                                          &                       & IC-Conv   &40.7$_{(\textbf{+1.0})}$	      \\ \hline
      \multirow{2}{*}{FCOS~\cite{tian2019fcos}}               & \multirow{2}{*}{ResNet50}  & Standard  &37.2	         \\
                                          &                       & IC-Conv   &38.8$_{(\textbf{+1.6})}$       \\ \hline
      \multirow{2}{*}{Faster R-CNN (NAS-FPN)~\cite{ghiasi2019fpn}} & \multirow{2}{*}{ResNet50}  & Standard  &40.2	       \\
                                          &                       & IC-Conv   &41.1$_{(\textbf{+0.9})}$   \\ 
      \bottomrule
      \end{tabular}}
      \end{center}
   \end{table}
   
\vspace{-2mm}
\begin{table*}[h]
\caption{Detection performance (AP\%) of different backbones when using  the standard convolution and our inception convolution on the COCO
2017 validation set.}
\label{tab:detector}
\label{tab1}
\begin{center}
\resizebox{400pt}{!}{%
\begin{tabular}{c|cc|llllll}
\toprule
Detector                      & Backbone                    & Conv Type & AP                       & \multicolumn{1}{c}{AP$_{50}$} & AP$_{75}$ & AP$_S$ & AP$_M$ & AP$_L$ \\ \hline
\multirow{6}{*}{Faster R-CNN~\cite{ren2015faster}}  & \multirow{2}{*}{ResNet50}        & Standard  & 36.4	    &58.6	&39.2	&21.7	&40.2	&46.4       \\
                              &                             & IC-Conv   & 38.9$_{(\textbf{+2.5})}$	&61.6	&41.8	&22.9	&42.3	&50.5       \\ \cline{2-9} 
                              & \multirow{2}{*}{ResNet101}       & Standard  & 38.8	    &60.9	&42.1	&22.6	&42.9	&50.5       \\
                              &                             & IC-Conv   & 41.9$_{(\textbf{+3.1})}$	&64.2	&45.3	&25.5	&45.8	&54.6       \\ \cline{2-9} 
                              & \multirow{2}{*}{ResNeXt101-32x4d} & Standard  & 40.5	    &63.1	&44.4	&24.9	&44.8	&52.0       \\
                              &                             & IC-Conv   & 42.1$_{(\textbf{+1.6})}$	&64.7	&45.7	&25.5	&46.1	&54.4       \\ \hline
\multirow{6}{*}{Cascade R-CNN~\cite{cai2018cascade}} & \multirow{2}{*}{ResNet50}        & Standard  & 40.5	    &59.2	&44.0	&22.5	&43.9	&53.6       \\
                              &                             & IC-Conv   & 42.4$_{(\textbf{+1.9})}$	&62.0	&46.0	&25.2	&45.9	&56.0       \\ \cline{2-9} 
                              & \multirow{2}{*}{ResNet101}       & Standard  & 42.6	    &60.9	&46.2	&23.8	&46.2	&56.9       \\
                              &                             & IC-Conv   & 45.0$_{(\textbf{+2.4})}$	&64.8	&48.7	&26.9	&49.0	&59.6       \\ \cline{2-9} 
                              & \multirow{2}{*}{ResNeXt101-32x4d} & Standard  & 44.4	    &63.6	&48.5	&25.8	&48.5	&58.1       \\
                              &                             & IC-Conv   & 45.7$_{(\textbf{+1.3})}$	&65.5	&49.7	&27.1	&49.8	&59.8       \\
\bottomrule
\end{tabular}}
\end{center}
\end{table*}
\vspace{-6mm}
\paragraph{Different detectors.}
We transfer our searched inception convolution to more types of detectors, including the one-stage detector RetinaNet~\cite{lin2017focal}, the anchor box free detector FCOS~\cite{tian2019fcos}, NAS-FPN~\cite{ghiasi2019fpn}, and the transformer based detector DETR~\cite{carion2020end}. The experimental
results on COCO are reported in Table~\ref{tab:different_detector}. Equipped with our inception convolution, the average AP gain for the five detectors is 1.8\% on COCO without additional FLOPs cost. In particular, our inception convolution is compatible with the new transformer based detector DETR. In combination with our inception convolution, DETR can achieve the AP of 40.7\% on COCO, which is 1.0\% higher than the vanilla DETR method. 

\vspace{-6mm}

\paragraph{Different datasets.} We also directly evaluate our searched inception convolution on another object detection dataset VOC~\cite{everingham2015pascal} by using the RFCN framework~\cite{dai2016r}. As shown in Table~\ref{tab:pascal_voc}, in terms of AP$_{50}$, our backbones IC-ResNet50 and IC-ResNet101
achieve the improvements by $1.94\%$ and $1.59\%$, when compared with the baseline backbones.  
\vspace{2mm}

\begin{table}[h]
   \caption{Detection performance ($AP_{50}$\%) on VOC test2007 by using the RFCN framework with the standard convolution and our inception convolution in the backbones ResNet50 and ResNet101.}
   \vspace{-4mm}
   \label{tab:pascal_voc}  
   \begin{center}
   \vspace{2mm}
   \resizebox{210pt}{!}{%
   \begin{tabular}{cc|cc}
      \toprule
         ResNet50   & IC-ResNet50 & ResNet101 & IC-ResNet101 \\ \hline
    79.66 & \textbf{81.60}  & 81.36   & \textbf{82.95}        \\ 
   \bottomrule
   \end{tabular}}
   \end{center} 
\end{table}

\vspace{-4mm}
\subsection{Instance Segmentation}
\subsubsection{Dataset and implementation details}
We further search for the dilation patterns for the instance segmentation task by using the Mask R-CNN~\cite{he2017mask} framework on the MS COCO~\cite{lin2014microsoft} dataset. The supernet with the  kernel size of 9 is also used as the pre-trained model to generate the inception convolutions. For Mask R-CNN training, we use the same training configurations as in Section~\ref{detetion_training_cfg}.

\vspace{-2mm}
\subsubsection{Experimental results}

We search for the dilation patterns based on Mask R-CNN~\cite{he2017mask} and transfer it to Cascade Mask R-CNN~\cite{cai2019cascade}. As shown in Table~\ref{tab:seg}, our searched inception convolution consistently boosts the instance segmentation accuracy. For Mask R-CNN with FPN, in terms of box AP, our backbones of IC-ResNet50, IC-ResNet101 and IC-ResNeXt101-32x4d outperform the baseline backbones by 2.8\%, 2.8\% and 2.0\% respectively. In particular, for the stronger Cascade Mask R-CNN with FPN, our backbone of IC-ResNeXt101-32x4d can further improve the box AP of the baseline backbone by 1.5\%.

\subsection{Comparison with other dilation search methods}
According to the results available in other works, we choose Faster R-CNN and Mask R-CNN as two examples for comparison. In Table~\ref{tab:sota_compare}, we compare our method with POD~\cite{peng2019pod}, NATS~\cite{peng2019efficient}, PSConv\footnote{Dilation patterns in PSConv are not searched but specifically designed.}~\cite{li2020psconv}  and CRNAS~\cite{liang2019computation}. 
Remarkably, our method achieves better performance than all the other baseline methods when using different backbones and detectors. We have a larger channel-wise search space and we also propose a more efficient search algorithm, which leads to the improved detection accuracy.
\vspace{2mm}
\begin{table}[h]
  \caption{Comparison (AP\%) between our inception convolution search method with other dilation search methods on the COCO 2017 validation set. 
   The results in the column ``Standard Conv" are copied from their original works. These results are comparable, which shows that we do not use any implementation trick. R50, R101 and X101-32x4d denote ResNet50, ResNet101 and ResNeXt101-32x4d, respectively.}
   
    \label{tab:sota_compare}
    \begin{center} 
   \resizebox{235pt}{!}{%
   \begin{tabular}{c|c|c|cc}
   \toprule
   \multicolumn{1}{l|}{Method}       & \multicolumn{1}{l|}{Backbone} & Conv type & Standard Conv & Searched Conv \\ \hline
   \multirow{14}{*}{Faster R-CNN~\cite{ren2015faster}} & \multirow{5}{*}{R50}          & POD       & 36.2     & 37.9   \\
                                     &                               & NATS      & 36.4     & 38.4   \\
                                     &                               & PSConv   & 36.4     & 38.4   \\
                                     &                               & CRNAS     & 36.4     & 38.3   \\
                                     &                               & \textbf{Ours}   & 36.4     & \textbf{38.9}   \\ \cline{2-5} 
                                     & \multirow{5}{*}{R101}         & POD       & 38.6     & 40.1   \\
                                     &                               & NATS      & 38.6     & 40.4   \\
                                     &                               & PSConv    & 38.5     & 40.9   \\
                                     &                               & CRNAS     & 38.6     & 40.2   \\
                                     &                               &  \textbf{Ours}   & 38.8     & \textbf{41.9}   \\ \cline{2-5} 
                                     & \multirow{4}{*}{X101-32x4d}   & NATS      & 40.5     & 41.6   \\
                                     &                               & CRNAS     & 40.6     & 41.5   \\
                                     &                               & PSConv    & 40.1     & 41.3   \\
                                     &                               &  \textbf{Ours}   & 40.5     & \textbf{42.1}   \\ \hline
   \multirow{9}{*}{Mask R-CNN~\cite{he2017mask}}    & \multirow{4}{*}{R50}          & NATS      & 37.5     & 39.3   \\
                                     &                               & PSConv    & 37.3     & 39.4   \\
                                     &                               & CRNAS     & 37.6     & 39.1   \\
                                     &                               &  \textbf{Ours}   & 37.2     & \textbf{40.0}     \\ \cline{2-5} 
                                     & \multirow{3}{*}{R101}         & PSConv    & 39.4     & 41.6   \\
                                     &                               & CRNAS     & 39.7     & 41.5   \\
                                     &                               &  \textbf{Ours}   & 39.8     & \textbf{42.6}   \\ \cline{2-5} 
                                     & \multirow{2}{*}{X101-32x4d}   & PSConv    & 41.1     & 42.4   \\
                                     &                               &  \textbf{Ours}   & 41.4     & \textbf{43.4}   \\ 
   \bottomrule
   \end{tabular}} 
\end{center}
   \end{table}

\begin{table*}[t]
   \caption{Detection and instance segmentation results of Mask R-CNN and Cascade Mask R-CNN on the COCO 2017 validation set. Box AP (\%) and mask AP (\%) of the bounding boxes and the segmentation results are reported, respectively. R50, R101 and X101-32x4d denote ResNet50, ResNet101 and ResNeXt101-32x4d, respectively.}
   \label{tab:seg}   
   \begin{center}
   \resizebox{495pt}{!}{%
   \begin{tabular}{c|c|c|llllll|llllll}
      \toprule
   \multirow{2}{*}{Detector} &
     \multirow{2}{*}{Backbone} &
     \multirow{2}{*}{Conv Type} &
     \multicolumn{6}{c|}{Box AP} &
     \multicolumn{6}{c}{Mask AP} \\ \cline{4-15} 
    &
      &
      &
     AP &
     \multicolumn{1}{c}{AP$_{50}$} &
     AP$_{75}$ &
     AP$_S$ &
     AP$_M$ &
     AP$_L$ &
     AP &
     \multicolumn{1}{c}{AP$_{50}$} &
     AP$_{75}$ &
     AP$_S$ &
     AP$_M$ &
     AP$_L$ \\ \hline
   \multirow{6}{*}{Mask R-CNN~\cite{he2017mask}} &
     \multirow{2}{*}{R50} &
     Standard &
     37.2 &
     59.0 &
     40.1 &
     22.3 &
     41.2 &
     47.7 &
     33.8 &
     55.7 &
     35.8 &
     17.9 &
     37.6 &
     45.9 \\
    &
      &
     IC-Conv &
     40.0$_{(\textbf{+2.8})}$ &
     62.1 &
     43.1 &
     23.5 &
     43.7 &
     52.1 &
     35.9$_{(\textbf{+1.9})}$ &
     58.4 &
     37.9 &
     18.9 &
     39.5 &
     49.5 \\ \cline{2-15} 
    &
     \multirow{2}{*}{R101} &
     Standard &
     39.8 &
     61.8 &
     43.5 &
     23.2 &
     43.9 &
     51.9 &
     35.6 &
     58.1 &
     38.1 &
     18.4 &
     39.5 &
     49.0 \\
    &
      &
     IC-Conv &
     42.6$_{(\textbf{+2.8})}$ &
     64.6 &
     46.4 &
     25.6 &
     46.6 &
     55.4 &
     37.9$_{(\textbf{+2.3})}$ &
     61.2 &
     40.3 &
     20.5 &
     41.7 &
     52.0 \\ \cline{2-15} 
    &
     \multirow{2}{*}{X101-32x4d} &
     Standard &
     41.4 &
     63.6 &
     45.2 &
     24.6 &
     45.9 &
     53.2 &
     37.1 &
     60.1 &
     39.6 &
     19.4 &
     41.3 &
     50.9 \\
    &
      &
     IC-Conv &
     43.4$_{(\textbf{+2.0})}$ &
     65.7 &
     47.4 &
     27.2 &
     47.2 &
     56.3 &
     38.4$_{(\textbf{+1.3})}$ &
     62.1 &
     40.4 &
     21.3 &
     42.0 &
     53.0 \\ \hline
   \multirow{6}{*}{Cascade Mask R-CNN~\cite{cai2019cascade}} & \multirow{2}{*}{R50} & Standard & 41.2 & 59.7 & 44.8 & 23.4 & 44.6 & 54.7 & 35.0 & 56.5 & 37.4 & 18.0 & 38.3 & 48.5 \\
    &
      &
     IC-Conv &
     43.4$_{(\textbf{+2.2})}$ &
     62.5 &
     47.0 &
     25.3 &
     47.1 &
     57.1 &
     36.8$_{(\textbf{+1.8})}$ &
     59.2 &
     39.2 &
     19.4 &
     40.1 &
     50.8 \\ \cline{2-15} 
    &
     \multirow{2}{*}{R101} &
     Standard &
     43.1 &
     61.9 &
     46.8 &
     24.8 &
     47.0 &
     56.7 &
     37.0 &
     59.0 &
     39.6 &
     19.0 &
     40.7 &
     50.8 \\
    &
      &
     IC-Conv &
     45.7$_{(\textbf{+2.6})}$ &
     65.2 &
     49.8 &
     26.8 &
     49.6 &
     61.0 &
     38.7$_{(\textbf{+1.7})}$ &
     61.9 &
     41.3 &
     20.5 &
     42.2 &
     53.9 \\ \cline{2-15} 
    &
     \multirow{2}{*}{X101-32x4d} &
     Standard &
     44.9 &
     64.1 &
     48.9 &
     26.1 &
     48.3 &
     59.4 &
     37.9 &
     60.6 &
     40.6 &
     20.0 &
     41.2 &
     52.6 \\
    &
      &
     IC-Conv &
     46.4$_{(\textbf{+1.5})}$ &
     66.0 &
     50.5 &
     27.1 &
     50.3 &
     61.0 &
     39.1$_{(\textbf{+1.2})}$ &
     62.6 &
     41.6 &
     20.6 &
     42.7 &
     53.7 \\ 
     \bottomrule
   \end{tabular}}
\end{center}
   \end{table*}

\subsection{Crowd Human Detection}
In crowd scenarios, different people have large variations in terms of poses and scales. An optimal receptive field need to be learned for correctly perceiving neighboring features. 

We use the CrowdHuman~\cite{shao2018crowdhuman} dataset, which contains 15K images for training.
Faster R-CNN with ResNet50 is adopted as the framework and the supernet kernel size is also 9. We use SGD as the optimizer with the momentum of 0.9 and the weight decay of 0.0001. The model is trained for 20 epochs and the learning rate is divided by 10 at the 10th and the 15th epochs.
As shown in Table~\ref{tab:crowdhuman}, on CrowdHuman, our backbone using inception convolution also outperforms the baseline backbone using the standard convolution by 0.8 \% in terms of $MR^{-2}$. Note $MR^{-2}$ is the most important metric on CrowdHuman and the results demonstrate the generalization capability of our inception convolution for different detection tasks. 

\begin{table}[h]
   \caption{Detection performance of Faster R-CNN (C4) on \textit{CrowdHuman} when using ResNet50 as the backbone together with the standard convolution and our inception convolution. 
   $MR^{-2}$ is the most critical indicator, in which lower value indicates better result.}
\label{tab:crowdhuman}   
\begin{center}
   \begin{tabular}{l|lll}
   \toprule
   Conv Type & Recall & AP    & $MR^{-2}$   \\ \hline
   Standard  & 79.29  & 75.61 & 59.60 \\ \hline
   IC-Conv   & \textbf{79.32}  & \textbf{75.68} & \textbf{58.82}      \\ 
   \bottomrule
   \end{tabular}
\end{center}
\end{table}

\vspace{-4mm}
\subsection{Human Pose Estimation}
\vspace{-2mm}
Human pose estimation aims to locate keypoints or parts of human (e.g. elbow, wrist, etc). To fully validate the effectiveness of our inception convolution, we choose a more difficult multi-person pose estimation problem, where all the human parts in an image should be detected and the keypoints of the same person should be associated.

\textbf{Dataset and implementation details.}
We train our model on the COCO train2017 dataset and evaluate our approach on the val2017 dataset. The supernet with the largest kernel size of 13 is used for searching the inception convolutions. 
Following the setting in MMPose (\url{https://github.com/open-mmlab/mmpose}), we adopt associative embedding~\cite{newell2017associative} for bottom-up human pose estimation and use the Adam optimizer. The initial learning rate is set as $1\mathrm{e}{-3}$,
and then drops to $1\mathrm{e}{-4}$ and $1\mathrm{e}{-5}$ 
at the $200$th and the $260$th epochs, respectively.
The training process is terminated within $300$ epochs.

\textbf{Experimental results.}
In Table~\ref{tab:human_pose}, we report the results of our inception convolution and standard convolution for the same input size (\ie, 640x640). Our backbones using inception convolution is more than \textbf{7\%} higher than the baseline backbones using the standard convolution in terms of AP, which is a \textbf{significant} improvement. 

\begin{table*}
\caption{Comparison between inception convolution and standard convolution for pose estimation on the COCO 2017 validation set.}
\label{tab:human_pose}
\begin{center}
\resizebox{380pt}{!}{%
\begin{tabular}{c|ccclccccc}
\toprule 
\multicolumn{1}{c|}{Method} &
  \multicolumn{1}{c|}{Backbone} &
  \multicolumn{1}{c|}{Input Size} &
  \multicolumn{1}{c|}{Conv Type} &
  AP &
  AP$_{50}$ & 
  AP$_{75}$ & 
  AP$_M$ & 
  AP$_L$ & 
  AR \\ \midrule
\multirow{10}{*}{\tabincell{c}{Associative\\ Embedding~\cite{newell2017associative}} } &
  \multicolumn{9}{c}{w/o multi-scale test setting} \\ \cmidrule{2-10} 
 &
  \multicolumn{1}{c|}{\multirow{2}{*}{ResNet50}} &
  \multicolumn{1}{c|}{\multirow{2}{*}{640x640}} &
  \multicolumn{1}{c|}{Standard} &
  51.0 &
  78.3 &
  52.3 &
  50.0 &
  52.0 &
  59.3 \\
 &
  \multicolumn{1}{c|}{} &
  \multicolumn{1}{c|}{} &
  \multicolumn{1}{c|}{IC-Conv} &
  62.2$_{(\textbf{+11.2})}$ &
  84.6 &
  67.8 &
  55.2 &
  72.2 &
  67.5 \\ \cmidrule{2-10} 
 & \multicolumn{1}{c|}{\multirow{2}{*}{ResNet101}} & \multicolumn{1}{c|}{\multirow{2}{*}{640x640}} & \multicolumn{1}{c|}{Standard} & 55.5 & 80.8 & 58.2 & 52.2 & 60.0   & 62.7 \\
 &
  \multicolumn{1}{c|}{} &
  \multicolumn{1}{c|}{} &
  \multicolumn{1}{c|}{IC-Conv} &
  63.3$_{(\textbf{+7.8})}$ &
  85.8 &
  69.4 &
  55.9 &
  74.0 &
  68.3 \\ \cmidrule{2-10} 
 &
  \multicolumn{9}{c}{w/ multi-scale test setting} \\ \cmidrule{2-10} 
 &
  \multicolumn{1}{c|}{\multirow{2}{*}{ResNet50}} &
  \multicolumn{1}{c|}{\multirow{2}{*}{640x640}} &
  \multicolumn{1}{c|}{Standard} &
  55.8 &
  81.0 &
  58.1 &
  56.3 &
  55.3 &
  63.8 \\
 &
  \multicolumn{1}{c|}{} &
  \multicolumn{1}{c|}{} &
  \multicolumn{1}{c|}{IC-Conv} &
  65.8$_{(\textbf{+10.0})}$ &
  85.9 &
  71.3 &
  60.3 &
  73.9 &
  70.7 \\ \cmidrule{2-10} 
 & \multicolumn{1}{c|}{\multirow{2}{*}{ResNet101}} & \multicolumn{1}{c|}{\multirow{2}{*}{640x640}} & \multicolumn{1}{c|}{Standard} & 60.2 & 83.8 & 63.3 & 58.7 & 62.4 & 67.2 \\
 &
  \multicolumn{1}{c|}{} &
  \multicolumn{1}{c|}{} &
  \multicolumn{1}{c|}{IC-Conv} &
  68.5$_{(\textbf{+8.3})}$ &
  87.9 &
  74.2 &
  63.6 &
  75.7 &
  73.0 \\ \bottomrule
\end{tabular}}
\end{center}
\end{table*}

 \vspace{-1mm}
\subsection{Analysis of Our Design}

\vspace{-1mm}
\subsubsection{Inception convolution and our EDO design}
\vspace{-2mm}
In this section, we take Faster R-CNN~\cite{ren2015faster} with FPN as an example to validate the effectiveness of both inception convolution and EDO.

\textbf{Enlarging the kernel size versus using inception convolution.} ResNet50 is the vanilla ResNet method. In ResNet50\_k9, we replace the 3×3 convolution operation in ResNet50 with the 9×9 convolution operation as our supernet also uses the kernel size of 9. In IC-ResNet50\_1d, we use the same dilation rate for both horizontal and vertical directions, thus we only have 1-dimensional freedom of dilation values. As shown in Table~\ref{tab:pattern}, based on the simplified 1-dimensional choice of dilation values, IC-ResNet50\_1d achieves a higher AP on COCO than the alternative method ResNet50\_k9. It indicates that our inception convolution can well capture the optimal receptive field for each channel. On the other hand, our complete method IC-ResNet50 enables searching for the optimal dilation values along both horizontal and vertical directions and thus outperforms IC-ResNet50\_1d as that searches for the optimal dilation patterns only along one direction.

\textbf{Searching Inception Convolution per layer or per channel.}
In IC-ResNet50\_sl, we keep the dilation values the same across all channels in each convolution layer. 

Our complete method IC-ResNet50 searches for the optimal dilation value for each channel, and it performs better than the alternative method IC-ResNet50\_sl that only conducts searching per layer.

\begin{table}[]
\caption{Detection performance (AP\%) of ResNet50, our method and different variants of our method on the COCO
2017 validation set.}
\label{tab:pattern}   
  \begin{center}
  \begin{tabular}{cc}
  \toprule
  Method & AP (\%)  \\ 
  \hline
  ResNet50           & 36.4 \\
  ResNet50\_k9       & 38.1 \\
  IC-ResNet50\_1d   & 38.3 \\
  IC-ResNet50\_sl   & 37.8 \\
  IC-ResNet50(Ours)               & \textbf{38.9} \\
  \bottomrule
  \end{tabular}
  \end{center}
\end{table}
\vspace{-2mm}
\subsubsection{Convolution kernel sizes of the supernet}
\vspace{-2mm}

\begin{figure}[] %[H]
    \centering 
    \includegraphics[width=0.28\textwidth]{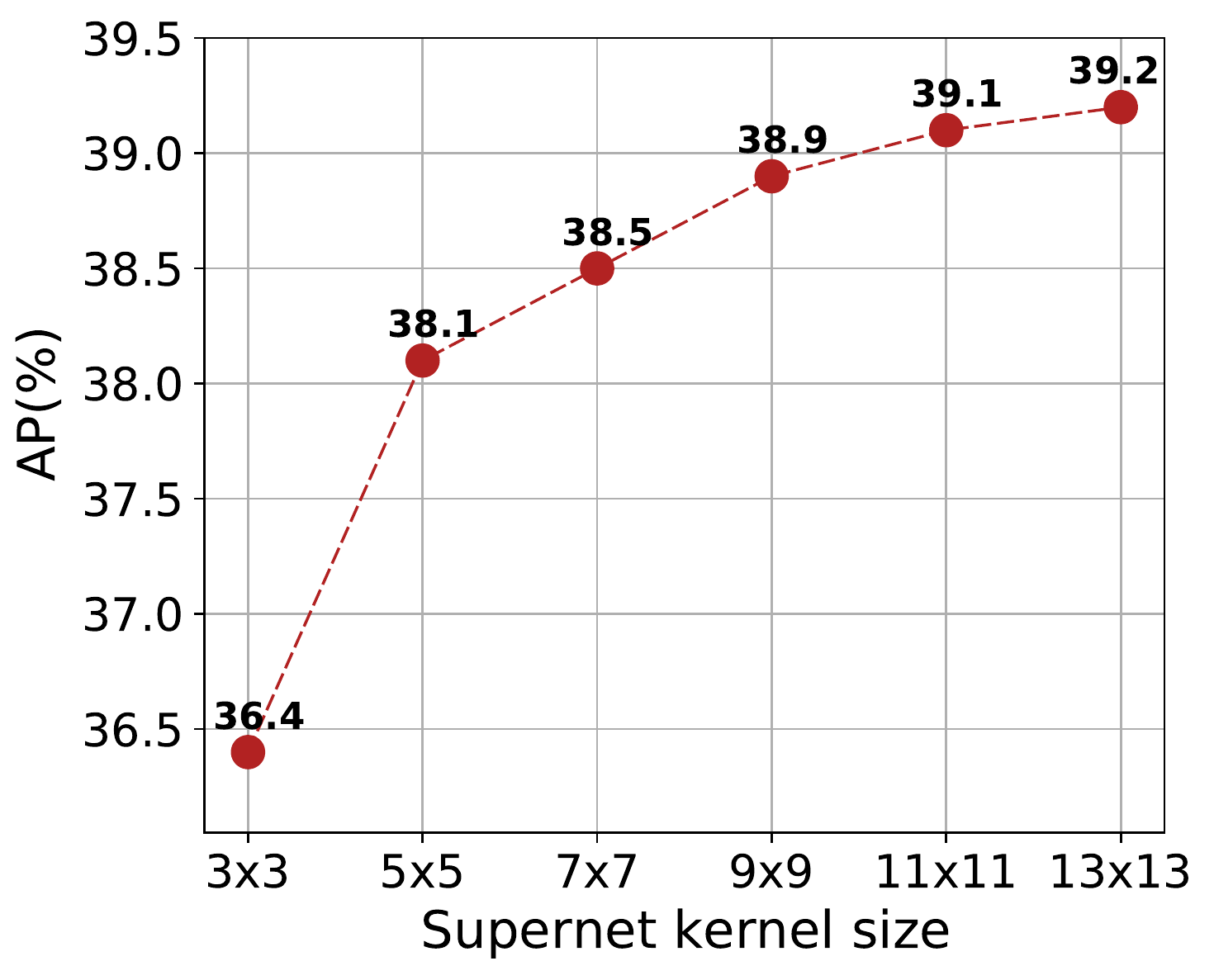}
    \caption{Detection performance (\%) of our method when using different supernet kernel sizes on the COCO
2017 validation set.}
    \label{fig:different_init_kernel}
\end{figure}

Larger kernel size introduces more dilation combination but brings more computation complexity in the retraining process of the supernet. 
In this section, we investigate the influence of the kernel size. 
We conduct all the detection experiments on COCO. As shown in Figure~\ref{fig:different_init_kernel}, 
 when the kernel size grows from 3x3 to 13x13, the AP grows consistently.
The result indicates that our method has great potential. By simply increasing the kernel size of the supernet from 9x9 to 13x13, higher gains can be achieved. By default, we set the kernel size as 9x9 because it is a good trade-off between accuracy and training time.
    
\vspace{-4mm} 

\subsubsection{Searching head/neck of the detector}
By default, we only search the optimal dilation patterns for the backbone part of the detector in Section~\ref{sec:object_detection}.
Here we further extend our method to search the head and neck parts of the detector. Specifically, we search all the 3×3 convolution in FPN~\cite{lin2017feature} and RPN~\cite{ren2015faster}. The results in Table~\ref{tab:neck_head} show that searching the dilation patterns for the head or neck part of the detector also leads to some improvement, though the improvement is less significant when compared with the improvement for the backbone.

\begin{table}[]
\caption{Searching different parts of the detector on the COCO
2017 validation set, in which we use Faster R-CNN with ResNet50 as the baseline.}
    \label{tab:neck_head}   
   \begin{center}
   \begin{tabular}{cccc}
   \toprule
   Backbone & FPN~\cite{lin2017feature} & RPN~\cite{ren2015faster} & AP (\%)  \\ \hline
   \checkmark        &     &     & 38.9 \\
   \checkmark        & \checkmark   &     & 39.0   \\
   \checkmark        & \checkmark   & \checkmark   & \textbf{39.2} \\ 
   \bottomrule
   \end{tabular}
   \end{center}
\end{table}

%% file: sections/5.conclusion.tex
\vspace{-0.1cm}
\section{Conclusion}
\vspace{-0.1cm}

In this work, we have proposed inception convolution, which is obtained by searching channel-wise dilation patterns through Efficient Dilation Optimization (EDO). IC-Conv can effectively decide the optimal receptive field in a convolution operation and aggregate the information induced by the receptive field at multiple scales. Our IC-Conv generalizes well for a large range of tasks and can be plugged into arbitrary CNN architectures.

\paragraph{Acknowledgments}
This work was supported by the National Key Research and Development Project of China (No. 2018AAA0101900). Wanli Ouyang was supported by the Australian Research Council Grant DP200103223 and Australian Medical Research Future Fund MRFAI000085.

%% file: cvpr.bbl
\begin{thebibliography}{10}\itemsep=-1pt

\bibitem{bender2018understanding}
Gabriel Bender, Pieter-Jan Kindermans, Barret Zoph, Vijay Vasudevan, and Quoc
  Le.
\newblock Understanding and simplifying one-shot architecture search.
\newblock In {\em International Conference on Machine Learning}, pages
  550--559, 2018.

\bibitem{cai2019once}
Han Cai, Chuang Gan, Tianzhe Wang, Zhekai Zhang, and Song Han.
\newblock Once-for-all: Train one network and specialize it for efficient
  deployment.
\newblock {\em arXiv preprint arXiv:1908.09791}, 2019.

\bibitem{cai2018proxylessnas}
Han Cai, Ligeng Zhu, and Song Han.
\newblock Proxylessnas: Direct neural architecture search on target task and
  hardware.
\newblock {\em arXiv preprint arXiv:1812.00332}, 2018.

\bibitem{cai2018cascade}
Zhaowei Cai and Nuno Vasconcelos.
\newblock Cascade r-cnn: Delving into high quality object detection.
\newblock In {\em Proceedings of the IEEE conference on computer vision and
  pattern recognition}, pages 6154--6162, 2018.

\bibitem{cai2019cascade}
Zhaowei Cai and Nuno Vasconcelos.
\newblock Cascade r-cnn: high quality object detection and instance
  segmentation.
\newblock {\em IEEE Transactions on Pattern Analysis and Machine Intelligence},
  2019.

\bibitem{carion2020end}
Nicolas Carion, Francisco Massa, Gabriel Synnaeve, Nicolas Usunier, Alexander
  Kirillov, and Sergey Zagoruyko.
\newblock End-to-end object detection with transformers.
\newblock {\em arXiv preprint arXiv:2005.12872}, 2020.

\bibitem{chen2017deeplab}
Liang-Chieh Chen, George Papandreou, Iasonas Kokkinos, Kevin Murphy, and Alan~L
  Yuille.
\newblock Deeplab: Semantic image segmentation with deep convolutional nets,
  atrous convolution, and fully connected crfs.
\newblock {\em IEEE transactions on pattern analysis and machine intelligence},
  40(4):834--848, 2017.

\bibitem{chen2017rethinking}
Liang-Chieh Chen, George Papandreou, Florian Schroff, and Hartwig Adam.
\newblock Rethinking atrous convolution for semantic image segmentation.
\newblock {\em arXiv preprint arXiv:1706.05587}, 2017.

\bibitem{dai2016r}
Jifeng Dai, Yi Li, Kaiming He, and Jian Sun.
\newblock R-fcn: Object detection via region-based fully convolutional
  networks.
\newblock In {\em Advances in neural information processing systems}, pages
  379--387, 2016.

\bibitem{dai2017deformable}
Jifeng Dai, Haozhi Qi, Yuwen Xiong, Yi Li, Guodong Zhang, Han Hu, and Yichen
  Wei.
\newblock Deformable convolutional networks.
\newblock In {\em Proceedings of the IEEE international conference on computer
  vision}, pages 764--773, 2017.

\bibitem{everingham2015pascal}
Mark Everingham, SM~Ali Eslami, Luc Van~Gool, Christopher~KI Williams, John
  Winn, and Andrew Zisserman.
\newblock The pascal visual object classes challenge: A retrospective.
\newblock {\em International journal of computer vision}, 111(1):98--136, 2015.

\bibitem{ghiasi2019fpn}
Golnaz Ghiasi, Tsung-Yi Lin, and Quoc~V Le.
\newblock Nas-fpn: Learning scalable feature pyramid architecture for object
  detection.
\newblock In {\em Proceedings of the IEEE conference on computer vision and
  pattern recognition}, pages 7036--7045, 2019.

\bibitem{Detectron2018}
Ross Girshick, Ilija Radosavovic, Georgia Gkioxari, Piotr Doll\'{a}r, and
  Kaiming He.
\newblock Detectron.
\newblock \url{https://github.com/facebookresearch/detectron}, 2018.

\bibitem{guo2020powering}
Ronghao Guo, Chen Lin, Chuming Li, Keyu Tian, Ming Sun, Lu Sheng, and Junjie
  Yan.
\newblock Powering one-shot topological nas with stabilized share-parameter
  proxy.
\newblock {\em arXiv preprint arXiv:2005.10511}, 2020.

\bibitem{guo2020single}
Zichao Guo, Xiangyu Zhang, Haoyuan Mu, Wen Heng, Zechun Liu, Yichen Wei, and
  Jian Sun.
\newblock Single path one-shot neural architecture search with uniform
  sampling.
\newblock In {\em European Conference on Computer Vision}, pages 544--560.
  Springer, 2020.

\bibitem{he2017mask}
Kaiming He, Georgia Gkioxari, Piotr Doll{\'a}r, and Ross Girshick.
\newblock Mask r-cnn.
\newblock In {\em Proceedings of the IEEE international conference on computer
  vision}, pages 2961--2969, 2017.

\bibitem{he2016deep}
Kaiming He, Xiangyu Zhang, Shaoqing Ren, and Jian Sun.
\newblock Deep residual learning for image recognition.
\newblock In {\em Proceedings of the IEEE conference on computer vision and
  pattern recognition}, pages 770--778, 2016.

\bibitem{krizhevsky2017imagenet}
Alex Krizhevsky, Ilya Sutskever, and Geoffrey~E Hinton.
\newblock Imagenet classification with deep convolutional neural networks.
\newblock {\em Communications of the ACM}, 60(6):84--90, 2017.

\bibitem{li2020psconv}
Duo Li, Anbang Yao, and Qifeng Chen.
\newblock Psconv: Squeezing feature pyramid into one compact poly-scale
  convolutional layer.
\newblock {\em arXiv preprint arXiv:2007.06191}, 2020.

\bibitem{li2020improving}
Xiang Li, Chen Lin, Chuming Li, Ming Sun, Wei Wu, Junjie Yan, and Wanli Ouyang.
\newblock Improving one-shot nas by suppressing the posterior fading.
\newblock In {\em Proceedings of the IEEE/CVF Conference on Computer Vision and
  Pattern Recognition}, pages 13836--13845, 2020.

\bibitem{li2019scale}
Yanghao Li, Yuntao Chen, Naiyan Wang, and Zhaoxiang Zhang.
\newblock Scale-aware trident networks for object detection.
\newblock In {\em Proceedings of the IEEE international conference on computer
  vision}, pages 6054--6063, 2019.

\bibitem{li2018detnet}
Zeming Li, Chao Peng, Gang Yu, Xiangyu Zhang, Yangdong Deng, and Jian Sun.
\newblock Detnet: A backbone network for object detection.
\newblock {\em arXiv preprint arXiv:1804.06215}, 2018.

\bibitem{liang2019computation}
Feng Liang, Chen Lin, Ronghao Guo, Ming Sun, Wei Wu, Junjie Yan, and Wanli
  Ouyang.
\newblock Computation reallocation for object detection.
\newblock {\em arXiv preprint arXiv:1912.11234}, 2019.

\bibitem{lin2017feature}
Tsung-Yi Lin, Piotr Doll{\'a}r, Ross Girshick, Kaiming He, Bharath Hariharan,
  and Serge Belongie.
\newblock Feature pyramid networks for object detection.
\newblock In {\em Proceedings of the IEEE conference on computer vision and
  pattern recognition}, pages 2117--2125, 2017.

\bibitem{lin2017focal}
Tsung-Yi Lin, Priya Goyal, Ross Girshick, Kaiming He, and Piotr Doll{\'a}r.
\newblock Focal loss for dense object detection.
\newblock In {\em Proceedings of the IEEE international conference on computer
  vision}, pages 2980--2988, 2017.

\bibitem{lin2014microsoft}
Tsung-Yi Lin, Michael Maire, Serge Belongie, James Hays, Pietro Perona, Deva
  Ramanan, Piotr Doll{\'a}r, and C~Lawrence Zitnick.
\newblock Microsoft coco: Common objects in context.
\newblock In {\em European conference on computer vision}, pages 740--755.
  Springer, 2014.

\bibitem{liu2018darts}
Hanxiao Liu, Karen Simonyan, and Yiming Yang.
\newblock Darts: Differentiable architecture search.
\newblock {\em arXiv preprint arXiv:1806.09055}, 2018.

\bibitem{liu2020block}
Jiaheng Liu, Shunfeng Zhou, Yichao Wu, Ken Chen, Wanli Ouyang, and Dong Xu.
\newblock Block proposal neural architecture search.
\newblock {\em IEEE Transactions on Image Processing}, 30:15--25, 2020.

\bibitem{luo2016understanding}
Wenjie Luo, Yujia Li, Raquel Urtasun, and Richard Zemel.
\newblock Understanding the effective receptive field in deep convolutional
  neural networks.
\newblock In {\em Advances in neural information processing systems}, pages
  4898--4906, 2016.

\bibitem{newell2017associative}
Alejandro Newell, Zhiao Huang, and Jia Deng.
\newblock Associative embedding: End-to-end learning for joint detection and
  grouping.
\newblock In {\em Advances in neural information processing systems}, pages
  2277--2287, 2017.

\bibitem{peng2019pod}
Junran Peng, Ming Sun, Zhaoxiang Zhang, Tieniu Tan, and Junjie Yan.
\newblock Pod: practical object detection with scale-sensitive network.
\newblock In {\em Proceedings of the IEEE International Conference on Computer
  Vision}, pages 9607--9616, 2019.

\bibitem{peng2019efficient}
Junran Peng, Ming Sun, ZHAO-XIANG ZHANG, Tieniu Tan, and Junjie Yan.
\newblock Efficient neural architecture transformation search in channel-level
  for object detection.
\newblock In {\em Advances in Neural Information Processing Systems}, pages
  14313--14322, 2019.

\bibitem{ren2015faster}
Shaoqing Ren, Kaiming He, Ross Girshick, and Jian Sun.
\newblock Faster r-cnn: Towards real-time object detection with region proposal
  networks.
\newblock In {\em Advances in neural information processing systems}, pages
  91--99, 2015.

\bibitem{russakovsky2015imagenet}
Olga Russakovsky, Jia Deng, Hao Su, Jonathan Krause, Sanjeev Satheesh, Sean Ma,
  Zhiheng Huang, Andrej Karpathy, Aditya Khosla, Michael Bernstein, et~al.
\newblock Imagenet large scale visual recognition challenge.
\newblock {\em International journal of computer vision}, 115(3):211--252,
  2015.

\bibitem{sandler2018mobilenetv2}
Mark Sandler, Andrew Howard, Menglong Zhu, Andrey Zhmoginov, and Liang-Chieh
  Chen.
\newblock Mobilenetv2: Inverted residuals and linear bottlenecks.
\newblock In {\em Proceedings of the IEEE conference on computer vision and
  pattern recognition}, pages 4510--4520, 2018.

\bibitem{shao2018crowdhuman}
Shuai Shao, Zijian Zhao, Boxun Li, Tete Xiao, Gang Yu, Xiangyu Zhang, and Jian
  Sun.
\newblock Crowdhuman: A benchmark for detecting human in a crowd.
\newblock {\em arXiv preprint arXiv:1805.00123}, 2018.

\bibitem{simonyan2014very}
Karen Simonyan and Andrew Zisserman.
\newblock Very deep convolutional networks for large-scale image recognition.
\newblock {\em arXiv preprint arXiv:1409.1556}, 2014.

\bibitem{szegedy2016inception}
Christian Szegedy, Sergey Ioffe, Vincent Vanhoucke, and Alex Alemi.
\newblock Inception-v4, inception-resnet and the impact of residual connections
  on learning.
\newblock {\em arXiv preprint arXiv:1602.07261}, 2016.

\bibitem{szegedy2015going}
Christian Szegedy, Wei Liu, Yangqing Jia, Pierre Sermanet, Scott Reed, Dragomir
  Anguelov, Dumitru Erhan, Vincent Vanhoucke, and Andrew Rabinovich.
\newblock Going deeper with convolutions.
\newblock In {\em Proceedings of the IEEE conference on computer vision and
  pattern recognition}, pages 1--9, 2015.

\bibitem{szegedy2016rethinking}
Christian Szegedy, Vincent Vanhoucke, Sergey Ioffe, Jon Shlens, and Zbigniew
  Wojna.
\newblock Rethinking the inception architecture for computer vision.
\newblock In {\em Proceedings of the IEEE conference on computer vision and
  pattern recognition}, pages 2818--2826, 2016.

\bibitem{tian2019fcos}
Zhi Tian, Chunhua Shen, Hao Chen, and Tong He.
\newblock Fcos: Fully convolutional one-stage object detection.
\newblock In {\em Proceedings of the IEEE international conference on computer
  vision}, pages 9627--9636, 2019.

\bibitem{xie2017aggregated}
Saining Xie, Ross Girshick, Piotr Doll{\'a}r, Zhuowen Tu, and Kaiming He.
\newblock Aggregated residual transformations for deep neural networks.
\newblock In {\em Proceedings of the IEEE conference on computer vision and
  pattern recognition}, pages 1492--1500, 2017.

\bibitem{yu2015multi}
Fisher Yu and Vladlen Koltun.
\newblock Multi-scale context aggregation by dilated convolutions.
\newblock {\em arXiv preprint arXiv:1511.07122}, 2015.

\bibitem{zela2019understanding}
Arber Zela, Thomas Elsken, Tonmoy Saikia, Yassine Marrakchi, Thomas Brox, and
  Frank Hutter.
\newblock Understanding and robustifying differentiable architecture search.
\newblock {\em arXiv preprint arXiv:1909.09656}, 2019.

\bibitem{zhang2017scale}
Rui Zhang, Sheng Tang, Yongdong Zhang, Jintao Li, and Shuicheng Yan.
\newblock Scale-adaptive convolutions for scene parsing.
\newblock In {\em Proceedings of the IEEE International Conference on Computer
  Vision}, pages 2031--2039, 2017.

\bibitem{zhao2017pyramid}
Hengshuang Zhao, Jianping Shi, Xiaojuan Qi, Xiaogang Wang, and Jiaya Jia.
\newblock Pyramid scene parsing network.
\newblock In {\em Proceedings of the IEEE conference on computer vision and
  pattern recognition}, pages 2881--2890, 2017.

\bibitem{zhou2020econas}
Dongzhan Zhou, Xinchi Zhou, Wenwei Zhang, Chen~Change Loy, Shuai Yi, Xuesen
  Zhang, and Wanli Ouyang.
\newblock Econas: Finding proxies for economical neural architecture search.
\newblock In {\em Proceedings of the IEEE/CVF Conference on Computer Vision and
  Pattern Recognition}, pages 11396--11404, 2020.

\bibitem{zoph2016neural}
Barret Zoph and Quoc~V Le.
\newblock Neural architecture search with reinforcement learning.
\newblock {\em arXiv preprint arXiv:1611.01578}, 2016.

\bibitem{zoph2018learning}
Barret Zoph, Vijay Vasudevan, Jonathon Shlens, and Quoc~V Le.
\newblock Learning transferable architectures for scalable image recognition.
\newblock In {\em Proceedings of the IEEE conference on computer vision and
  pattern recognition}, pages 8697--8710, 2018.

\end{thebibliography}
